\documentclass[letterpaper]{article} %
\usepackage{aaai2027}  %
\usepackage[hyphens]{url}  %
\usepackage{graphicx} %
\usepackage{natbib}  %
\usepackage{caption} %
\usepackage{newfloat}
\usepackage{listings}
\DeclareCaptionStyle{ruled}{labelfont=normalfont,labelsep=colon,strut=off} %
\usepackage{tabularx}
\newcolumntype{C}{>{\centering\arraybackslash}X}
\usepackage{booktabs}
\definecolor{rankone}{RGB}{255, 210, 210}   %
\definecolor{ranktwo}{RGB}{255, 225, 190}   %
\definecolor{rankthree}{RGB}{255, 245, 200} %
\newcommand{\first}[1]{\cellcolor{rankone}\textbf{#1}}
\newcommand{\second}[1]{\cellcolor{ranktwo}#1}
\newcommand{\third}[1]{\cellcolor{rankthree}#1}

\newcommand\ourme{TRACE\xspace}
\newcommand\baselineAR{ActiveGS-Random\xspace}
\newcommand\baselineAU{ActiveGS-Uniform\xspace}

\usepackage{amsmath, amssymb, mathtools}
\usepackage{amsthm}
\theoremstyle{definition}
\newtheorem{definition}{Definition}
\newtheorem{problem}{Problem}
\theoremstyle{plain}
\newtheorem{proposition}{Proposition}
\usepackage{url}
\usepackage[ruled,linesnumbered]{algorithm2e}   %
\usepackage{graphicx}                            %
\usepackage{color}
\usepackage{booktabs}
\usepackage{multirow}
\usepackage{makecell}
\usepackage[table]{xcolor}
\usepackage{algorithmic}

\definecolor{cvprblue}{rgb}{0.21,0.49,0.74}
\usepackage[
    colorlinks=true,
    breaklinks=true,     
    linkcolor=black,    
    citecolor=black,     
    filecolor=black,
    urlcolor=cvprblue     
]{hyperref}

\usepackage{cleveref}

\title{TRACE: Ergodic Trajectory Optimization for Active Scene Reconstruction}

\author {
    Ziyue Zheng\textsuperscript{\rm 1}\equalcontrib,
    Linli Shi\textsuperscript{\rm 1}\equalcontrib,
    Bingkun He\textsuperscript{\rm 1},
    Wen Jiang\textsuperscript{\rm 2},
    Ziyun Wang\textsuperscript{\rm 1}\corresponding
}
\affiliations {
    \textsuperscript{\rm 1}Johns Hopkins University, USA\\
    \textsuperscript{\rm 2}University of Pennsylvania, USA\\
    \{zzheng60, lshi42, bhe15\}@jh.edu, wenjiang@seas.upenn.edu, claude.w@jhu.edu\\
    {\footnotesize\textsuperscript{*}These authors contributed equally.\quad
     \textsuperscript{\dag}Corresponding author.}
}

\begin{document}
\twocolumn[{%
\centering
\renewcommand\twocolumn[1][]{#1}%
\maketitle
\vspace{-8mm}
\includegraphics[width=0.96\linewidth]{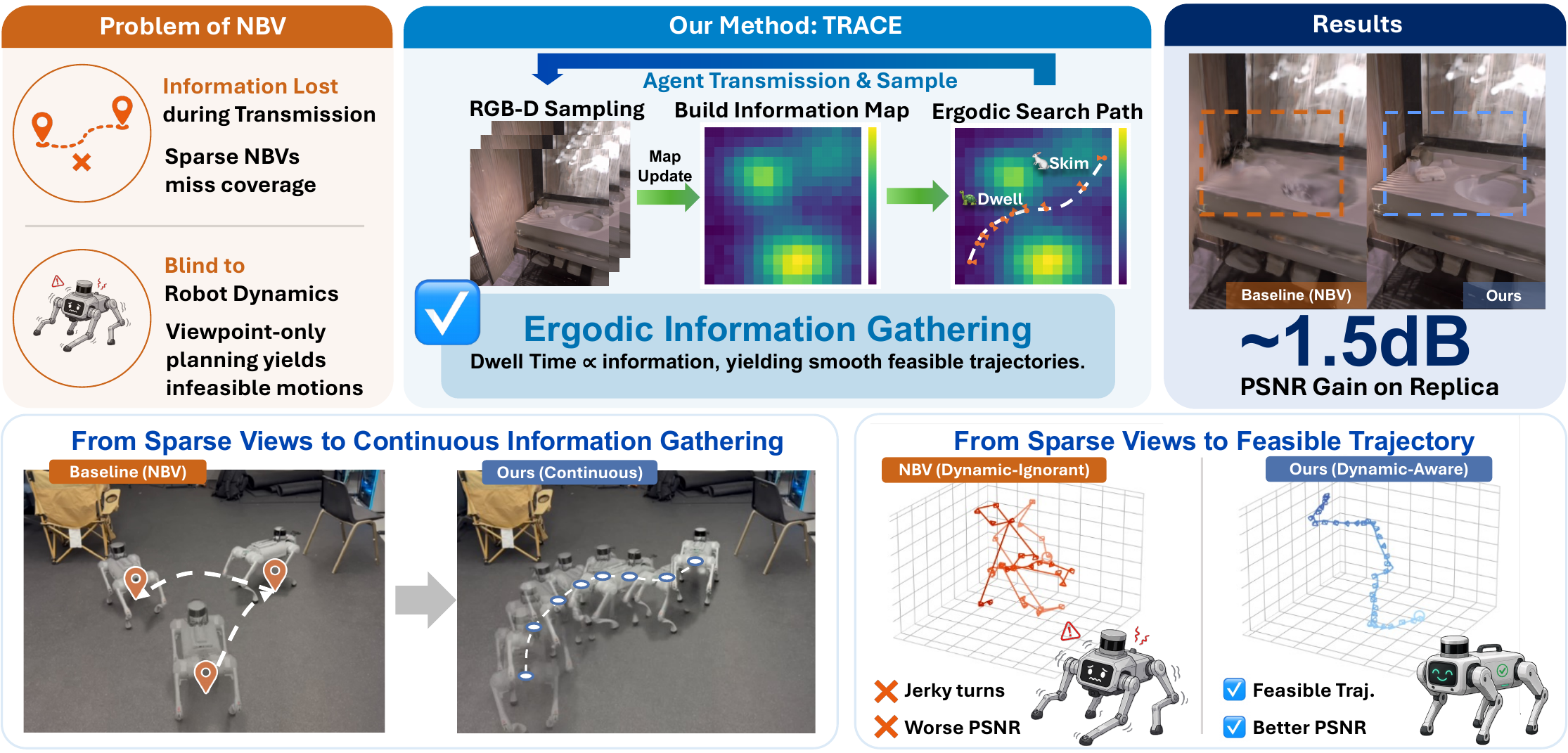}
\captionsetup{hypcap=false}
\captionof{figure}{\textbf{From discrete viewpoints to ergodic active reconstruction.}
    Next-Best-View (NBV) planners discard information between sparse viewpoints and ignore robot dynamics, producing jerky, infeasible trajectories. Our method instead constructs an information map from streaming RGB-D observations and plans an ergodic search path that continuously gathers information while respecting the robot's dynamic. This yields smoother trajectories, denser coverage, and 1.5 dB PSNR gain.
    }
    \vspace{6mm}
\label{fig:teaser}
}]

\begin{abstract}
Existing active reconstruction systems with Gaussian-splatting maps select observations greedily, optimizing a single next-best-view (NBV) at each step and connecting the chosen views by short-horizon path planning. This greedy decoupling disregards the global structure of scene information, producing inefficient trajectories that waste sensing capacity in transit between selected views. In this work, we study active reconstruction as an ergodic coverage problem: the time-averaged spatial statistics of the sensor trajectory should match a target information distribution induced by the current map. Our approach derives this target distribution online from 
uncertainty and visibility, and calculates ergodic trajectories via a kernel-ergodic horizon planner with gradient flow and footprint depletion, closing the loop between mapping and trajectory optimization. We thoroughly evaluate \ourme on the Replica dataset against the Next-Best-View (NBV) baselines, improving PSNR by 1.5~dB. \textbf{Code}: 
\href{https://github.com/spikelab-jhu/trace-active-reconstruction}{https://github.com/spikelab-jhu/trace-active-reconstruction}.

\end{abstract}

\section{Introduction}\label{sec:intro}
Active 3D reconstruction recovers the geometry of an unknown scene by moving a sensor along a sequence of informative viewpoints, which has long been a core problem in robotic perception~\citep{10897340, Isler, bircher2016receding}. In such tasks, the robot incrementally builds a high-fidelity model of the scene from its own streaming observations. Recent advances in differentiable rendering, particularly 3D Gaussian Splatting (GS)~\citep{kerbl2023gaussians} and its surface-aligned variant 2DGS~\citep{huang2024twodgs}, have raised the achievable reconstruction quality dramatically. The bottleneck has shifted from the map representation to the planner~\citep{jin2025activegs,activegamer, Li_2025}: under finite budgets of time, energy, and motion, the agent must decide how to spend each meter of motion across an unknown scene. Therefore, the reconstruction quality depends on both map representations and how the agent chooses to distribute its limited motion budget across the scene.

A common paradigm of active reconstruction with Gaussian-splatting maps is to plan view-by-view: at each step, the planner scores candidate viewpoints by an information-gain proxy, commits to the highest-scoring pose, and connects it to the current location by short-horizon path planning~\citep{jin2025activegs,activegamer}. Recent work introduces a hierarchy over a topological subgraph, but each step still commits to a single node and routes to it via shortest-path planning~\citep{Li_2025}. However, this formulation decouples viewpoint selection from trajectory planning, and the motion between successive views serves only to move the sensor toward the next target rather than to acquire additional information. The decoupling has \textit{three main issues}. First, viewpoint scoring is \textbf{short-term}: each decision commits to the locally most informative pose without anticipating where the sensor must subsequently travel. Second, the behaviors between selected views do not contribute to \textbf{efficient information gathering}. The resulting trajectories are jagged, redundant, and biased toward isolated high-information viewpoints rather than uniform spatial coverage of the scene. Third, the trajectory formed by the sequence of planned views ignores \textbf{real-world constraints} such as controllability, actuator limits, and energy efficiency, often producing dynamically infeasible trajectories. In particular, discrete search algorithms used in Next-Best-View (NBV) planning typically optimize over viewpoints rather than executable motions, and therefore cannot directly account for the robot’s continuous-time dynamics.

In this work, we aim to address these issues by proposing the first active 3D reconstruction method that plans continuous and control-feasible trajectories using ergodic search.
Instead of modeling the problem as a planning over a set of next poses, we directly optimize a trajectory over the constructed information distribution based on the current state of mapping and exploration.
Specifically, our planner optimizes the path so that its time-averaged spatial statistics match a target information distribution: the time the trajectory spends in any region is proportional to the information density there. 
Instantiating this formulation on a live Gaussian-splatting map raises
three technical challenges. First, classical ergodic search assumes a
target distribution specified a priori, whereas in active reconstruction
the information density is implicit in the map state, encoded in
per-Gaussian confidence and evolving occupancy, and must be re-derived
online as observations are integrated. Second, the informative regions
are scene surfaces that the sensor cannot occupy: the trajectory must
remain in free space while its viewing footprint, rather than its
position, covers the high-information surfaces. Third, the marginal
value of observing a surface diminishes under repeated coverage within a
horizon, a time-varying effect that a static target distribution cannot
capture.

The target information distribution is derived online from per-Gaussian
uncertainty and visibility, while the trajectory is computed by
a kernel-ergodic horizon planner with gradient-flow updates and a
footprint-depletion mechanism that suppresses re-coverage. Our contributions are summarized as:
\begin{itemize}
    \item To the best of our knowledge, we are the first to formulate active reconstruction with Gaussian-splatting maps as a trajectory-level ergodic coverage problem, replacing greedy viewpoint selection with continuous optimization over dynamically feasible trajectories.

    \item We adapt kernel-based ergodic search to surface-based reconstruction by diffusing surface information into traversable free space and introducing footprint-aware depletion and gaze objectives for efficient coverage and camera orientation.

    \item Under the same mapper, budget, and evaluation protocol, \ourme{} improves over the strongest NBV baseline by $1.5$~dB PSNR across eight Replica scenes and supports direct trajectory execution on quadruped and manipulator (supplementary) without an intermediate path planner. 
\end{itemize}

\section{Related Work}

 \subsection{Active 3D Reconstruction}
  \label{sec:related-active-recon}
  
  Active 3D reconstruction has been studied for decades under the
  \emph{next-best-view} (NBV) formulation~\citep{1087372,
  Isler,bircher2016receding,delmerico2018comparison}: at each step, the agent selects the
  viewpoint expected to maximize a chosen information-gain criterion.
  Sampling-based informative path planning extends NBV to entire
  trajectories~\citep{hollinger2014sampling} and long-horizon tree search~\citep{best2019decmcts}, while multi-stage aerial pipelines couple it to
  multi-view-stereo reconstruction~\citep{hepp2018plan3d}. Frontier-based exploration~\citep{yamauchi1997frontier} and occupancy
  mapping~\citep{octomap} provide complementary geometric
  drivers, and SCONE~\citep{guedon2022scone} optimizes surface
  coverage via Monte Carlo volumetric integration.

  The emergence of differentiable rendering---%
  NeRF~\citep{mildenhall2020nerf} and subsequent radiance-field
  variants~\citep{barron2022mipnerf360,muller2022instant,
  fridovichkeil2022plenoxels,chen2022tensorf}---extended active
  reconstruction to learned neural map representations.
  NARUTO~\citep{feng2024naruto}, ActiveNeRF~\citep{pan2022activenerf},
  ActiveImplicit~\citep{yan2023activeimplicit}, and
  NeU-NBV~\citep{jin2023neunbv} score viewpoints by NeRF uncertainty
  or implicit-occupancy entropy. FisherRF~\citep{jiang2024fisherrf}
  uses Fisher information over radiance-field parameters;
  NVF~\citep{xue2024nvf} composites position-based uncertainty into
  camera-ray uncertainty; Active Neural
  Mapping~\citep{yan2023activeneuralmap} measures neural variability
  under weight perturbation; and~\citet{he2024activeperception}
  maximize mutual information through a generative NeRF model.
  GenNBV~\citep{chen2024gennbv} trains a generalizable RL policy over
  a 5-DoF action space, while
  ACE-NBV~\citep{zhang2023affordancenbv} selects views that improve
  grasp quality. 

  Recent Gaussian-splatting maps~\citep{kerbl2023gaussians,
  huang2024twodgs} have prompted dedicated active-GS planners.
  ActiveGS~\citep{jin2025activegs},
  ActiveGAMER~\citep{activegamer}, and
  ActiveSplat~\citep{Li_2025} score viewpoints by per-primitive
  confidence, rendering-based information gain, or frontier cues;
  ActiveSplat further introduces a local-vs-global hierarchy over
  a topological subgraph. GauSS-MI~\citep{xie2025gaussmi} introduces
  Shannon mutual information over Gaussian-splat appearance,
  POp-GS~\citep{wilson2025popgs} reframes information gain through
  P-Optimality, and Active3D~\citep{li2025active3dactivehighfidelity3d}
  fuses implicit and explicit representations with hierarchical uncertainty
  quantification. These methods share a common structure: at each
  step, the planner commits to a best viewpoint and connects
  it to the current pose by short-horizon path planning, producing
  trajectories at the viewpoint level rather than the
  trajectory level.
\subsection{Ergodic Coverage for Robotic Exploration}
Ergodic search formulates information gathering as a trajectory-level 
coverage problem: the time-averaged spatial statistics of the trajectory should match a target information distribution~\citep{mathew2011metrics,miller2015ergodic}. 
This trajectory-centric formulation naturally accommodates finite sensing budgets by optimizing information collection over the entire motion rather than selecting isolated viewpoints.
To optimize the ergodic problem, a range of methods optimize the ergodic objective via spectral multi-scale coverage with Fourier analysis~\citep{mathew2011metrics}, Kullback–Leibler divergence~\citep{abraham2021ergodic}, kernel-based ergodic metrics~\citep{kenerl}, flow matching ~\citep{sun2025flowmatchingergodiccoverage}, receding-horizon control~\cite{mavrommati2017real}, potential field approaches~\cite{ivic2016ergodicity}, and LQR-based optimization~\cite{miller2013trajectory}. Subsequent work has extended the framework along several axes, including time-optimal
  ergodic search~\citep{dong2024time}, dynamic sensor
  footprints~\citep{zheng2025footprint}, probabilistic connectivity~\cite{2025_RSS_IMEC_YongceLiu}, manipulation~\cite{shetty2021ergodic}, target localization~\cite{mavrommati2017real} and coverage in constrained
  domains~\citep{ayvali2017ergodic}.
  Despite these advances, most of these methods typically assume the
  target distribution is parametrically given a priori from a model of
  expected information density. In contrast, we derive the target
  distribution online from a live 2DGS map, where the information density
  is implicit in per-Gaussian uncertainty and evolving visibility
  geometry. 
  To our knowledge, this is the first ergodic trajectory optimization
  formulation for active reconstruction with Gaussian-splatting maps.

 \section{Method}
  \label{sec:method}
  We instantiate the formulation of Sec.~\ref{sec:intro} as a horizon-based
  ergodic trajectory optimization driven by online 2DGS map. Sec.~\ref{sec:method-problem} formalizes the problem; Sec.~\ref{sec:method-phi} constructs the target information distribution from the map state; Sec.~\ref{sec:method-ergodic} optimizes a
  kernel-ergodic trajectory by gradient descent on a footprint-depleting information field.
  \subsection{Problem Formulation}
  \label{sec:method-problem}

  We consider the problem of active 2D Gaussian-surfel (2DGS) mapping: an
  autonomous agent equipped with an RGB-D sensor incrementally builds a 2DGS
  map $\mathcal{M}$ of an unknown bounded scene
  $\Omega \subset \mathbb{R}^3$ through a sequence of self-selected viewpoints.
  We adopt the 2DGS representation for its surface-aligned geometry, which
  directly supports the mesh-quality metrics commonly used to evaluate
  active reconstruction~\citep{huang2024twodgs}. The sensor is modeled as a view cone, rotationally symmetric about its optical axis, so its roll is immaterial and held fixed. At each replanning step
(horizon) $t$, the planner selects a $K$-step trajectory
$\tau_t = (\mathbf{p}_t^1, \dots, \mathbf{p}_t^K) \in SE(3)^K$, with positions
$\mathbf{x}_t^k \in \mathbb{R}^3$ and orientations determined by yaw and pitch.
The trajectory is parameterized by control
$\mathbf{u}_t \in \mathbb{R}^{K \times 5}$ (position velocity, yaw rate, pitch
rate) cumulatively integrated under single-integrator dynamics (matching Go2's interface). The supplementary shows a more complex case on FR3.

\begin{figure}[t]
    \centering
    \includegraphics[width=\columnwidth]
    {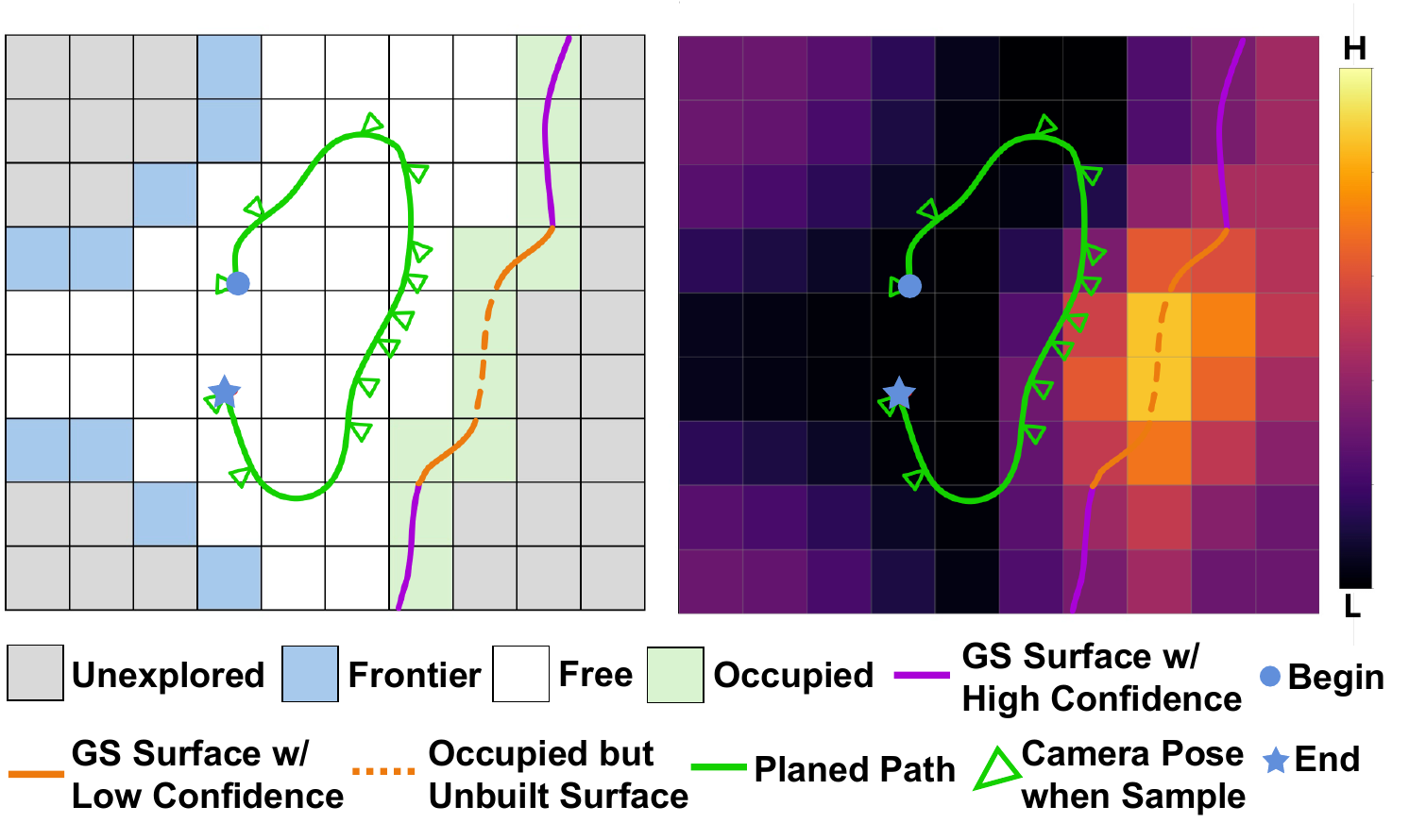}
    
    \caption{
    \textbf{2D illustration of a single planning horizon.} \textbf{Left:} voxel classification and our planned path. \textbf{Right:} the corresponding information-map heatmap (Sec.~\ref{sec:method-phi}). The planner jointly optimizes the trajectory, allocating sampling effort proportional to the information distribution, dwelling longer and sampling more densely in high-information regions.
    }
    \label{fig:infomap}
\end{figure}

  After executing the full horizon, the agent integrates new RGB-D
  observations into $\mathcal{M}_t$ and re-plans with $\mathbf{u}$
  warm-started from the previous optimization. 
  Although the voxel map is updated during execution, the planning objective remains fixed until the next horizon.
  Formally, each replanning step solves
  \begin{equation}
      \tau^\star_t \;=\; \arg\min_{\tau_t \in \mathcal{T}_t} \; J_t(\tau_t;\, \mathcal{M}_t,\mathcal{V}_t),
      \label{eq:objective}
  \end{equation}
where $\mathcal{T}_t$ is the feasible-trajectory set in horizon $t$
(collision-free, bounded step size) and $J_t$ is the ergodic optimal-control
objective of Sec.~\ref{sec:method-ergodic}, defined over a target information
distribution $\phi_t$ derived online from the current map
(Sec.~\ref{sec:method-phi}). In practice we relax the hard constraints defining
$\mathcal{T}_t$ into the soft penalties of Eq.~\eqref{eq:full-cost} and optimize
the resulting objective over the controls.

  \subsection{Information Map}\label{sec:method-phi}
  Ergodic search requires a target information distribution to guide trajectory planning. In our task, $\phi_t$ (information map at horizon $t$), evolves as new observations are integrated.
    To introduce the information map, we first present map representation design. We follow the hybrid map representation of ActiveGS~\cite{jin2025activegs}: a 2DGS Map $\mathcal{M}_t$ is used for rendering, while a voxel map $\mathcal{V}_t$ represents the occupancy probability. GS Map $\mathcal{M}_t$ is trained and updated at the end of each horizon, whereas the voxel map $\mathcal{V}_t$ is updated on the fly following the approach of OctoMap~\cite{octomap}. 

    For brevity we suppress the dependence on $(\mathcal{M}_t,\mathcal{V}_t)$ in the notation below; all fields are evaluated at the current map state. For each voxel $v$ in $\mathcal{V}_t$, the raw information value is

\begin{align}
\phi_t^{\mathrm{raw}}(v) \;&=\;
    \alpha_{\mathrm{u}}\, \mathbf{1}_{\mathrm{unexp}}(v)
    \;+\; \alpha_{\mathrm{f}}\, \mathbf{1}_{\mathrm{front}}(v)
    \;+\; \notag \\
    & \alpha_{\mathrm{b}}\, \mathbf{1}_{\mathrm{unbuilt}}(v)
    \;+\; \beta\,\bigl(1 - c(v)\bigr)\,\mathbf{1}_{\mathrm{low}}(v),
    \label{eq:phi-raw}
\end{align}

Here $\mathbf{1}_{\mathrm{unexp}}(v)$, $\mathbf{1}_{\mathrm{front}}(v)$, and
$\mathbf{1}_{\mathrm{unbuilt}}(v)$ indicate, respectively, voxels no depth ray has
traversed, free voxels bordering unexplored space, and voxels $\mathcal{V}_t$
deems occupied but holding no Gaussian of $\mathcal{M}_t$. 
Since $\mathcal{V}_t$ updates on execution but $\mathcal{M}_t$ only at horizon
boundaries, known-occupied voxels may still hold no Gaussian; $\mathbf{1}_{\mathrm{unbuilt}}$ draws the sensor back to them and reduces
holes in the GS map early in the mission.
For the confidence term, $\mathcal{G}_v \subseteq \mathcal{M}_t$
collects the Gaussians whose centers fall in voxel $v$, $\gamma_g \in [0,1]$ is the
per-primitive rendering confidence inherited from ActiveGS~\cite{jin2025activegs},
and $c(v)=\tfrac{1}{|\mathcal{G}_v|}\sum_{g\in\mathcal{G}_v}\gamma_g$ is their mean.
Intuitively, $\gamma_g$ grows with the number and angular diversity of viewpoints that have already observed Gaussian $g$, so $1-c(v)$ is large precisely on voxels whose surface fit is still weakly constrained. The gate $\mathbf{1}_{\mathrm{low}}(v)=1$ on voxels
that contain Gaussians but are not yet \emph{well built} (as shown in Fig.~\ref{fig:ergo_search_object}); well-built voxels contribute no
mass and stop attracting the sensor.
The non-negative weights $\alpha_{\mathrm{u}},\alpha_{\mathrm{f}},\alpha_{\mathrm{b}},\beta$ control, respectively, exploration of unseen volume, expansion of the frontier, attention to visible-but-unmodeled surfaces, and refinement of low-confidence Gaussians. 

\begin{table*}[ht]
\centering
\caption{Quantitative comparison of rendering quality on the Replica dataset.
We report PSNR, SSIM, and LPIPS across all scenes.
Best, second-best, and third-best results are highlighted in
\colorbox{rankone}{red}, \colorbox{ranktwo}{orange}, and
\colorbox{rankthree}{yellow}, respectively.}
\label{tab:replica_comparison}
\renewcommand{\arraystretch}{1}

\begin{tabularx}{\textwidth}{l*{9}{C}}
\toprule
\textbf{Methods} & \textbf{Metrics} & \textbf{Of0} & \textbf{Of2} & \textbf{Of3} & \textbf{Of4} & \textbf{R0} & \textbf{R1} & \textbf{R2} & \textbf{H0} \\
\midrule
NARUTO      & PSNR $\uparrow$ &       32.99  &       27.60  &       27.48  &       30.28  & {28.41} &       26.77  &       29.75  &       25.49  \\
FisherRF
& PSNR $\uparrow$ & \third{36.21} & \third{30.33} & \third{28.63} & \third{32.80} & \third{28.88} & \third{30.04} & \third{32.26} & \third{26.79} \\
ActiveGS
& PSNR $\uparrow$ & \second{36.78} & \second{31.68} & \second{32.16} & \second{34.08} & \second{29.93} & \second{31.74} & \second{32.35} & \second{32.04} \\
\textbf{Ours}                     & PSNR $\uparrow$ & \first{38.36} & \first{32.81} & \first{33.79} & \first{34.92} & \first{31.55} & \first{32.84} & \first{34.70} & \first{33.69} \\
\midrule
ActiveGS& SSIM $\uparrow$    & \second{0.963} & \second{0.926}  & \second{0.921} & \second{0.932} & \second{0.891} & \second{0.902} & \second{0.924} & \second{0.937} \\
\textbf{Ours}                     & SSIM $\uparrow$    & \first{0.969}  & \first{0.930} & \first{0.929}  & \first{0.934}  & \first{0.907}  & \first{0.914}  & \first{0.941}  & \first{0.953}  \\
ActiveGS
& LPIPS $\downarrow$ & \second{0.081} & \second{0.136} & \second{0.152} & \second{0.138} & \second{0.184} & \second{0.170} & \second{0.146} & \second{0.139} \\
\textbf{Ours}                     & LPIPS $\downarrow$ & \first{0.066}  & \first{0.117}  & \first{0.143}  & \first{0.122}  & \first{0.158}  & \first{0.151}  & \first{0.118}  & \first{0.110}  \\
\bottomrule

 \end{tabularx}
\end{table*}

We obtain $\phi_t$ by box-filtering $\phi_t^{\mathrm{raw}}$, masking it to the observed collision-free region of
$\mathcal{V}_t$, and normalizing to a probability distribution. Masking is what
makes the ergodic target realizable: it projects information mass from occupied
surfaces onto the reachable free space the sensor can actually occupy. The gaze
reward instead uses the unmasked $\phi_t^{\mathrm{raw}}$ to preserve absolute
scale. Both fields are sampled at continuous locations by trilinear
interpolation, so all terms are differentiable in the trajectory.

\subsection{Kernel-Ergodic Trajectory with Sensor Footprint Depletion}
  \label{sec:method-ergodic}

\begin{figure}
    \centering
    \includegraphics[width=.905\columnwidth]{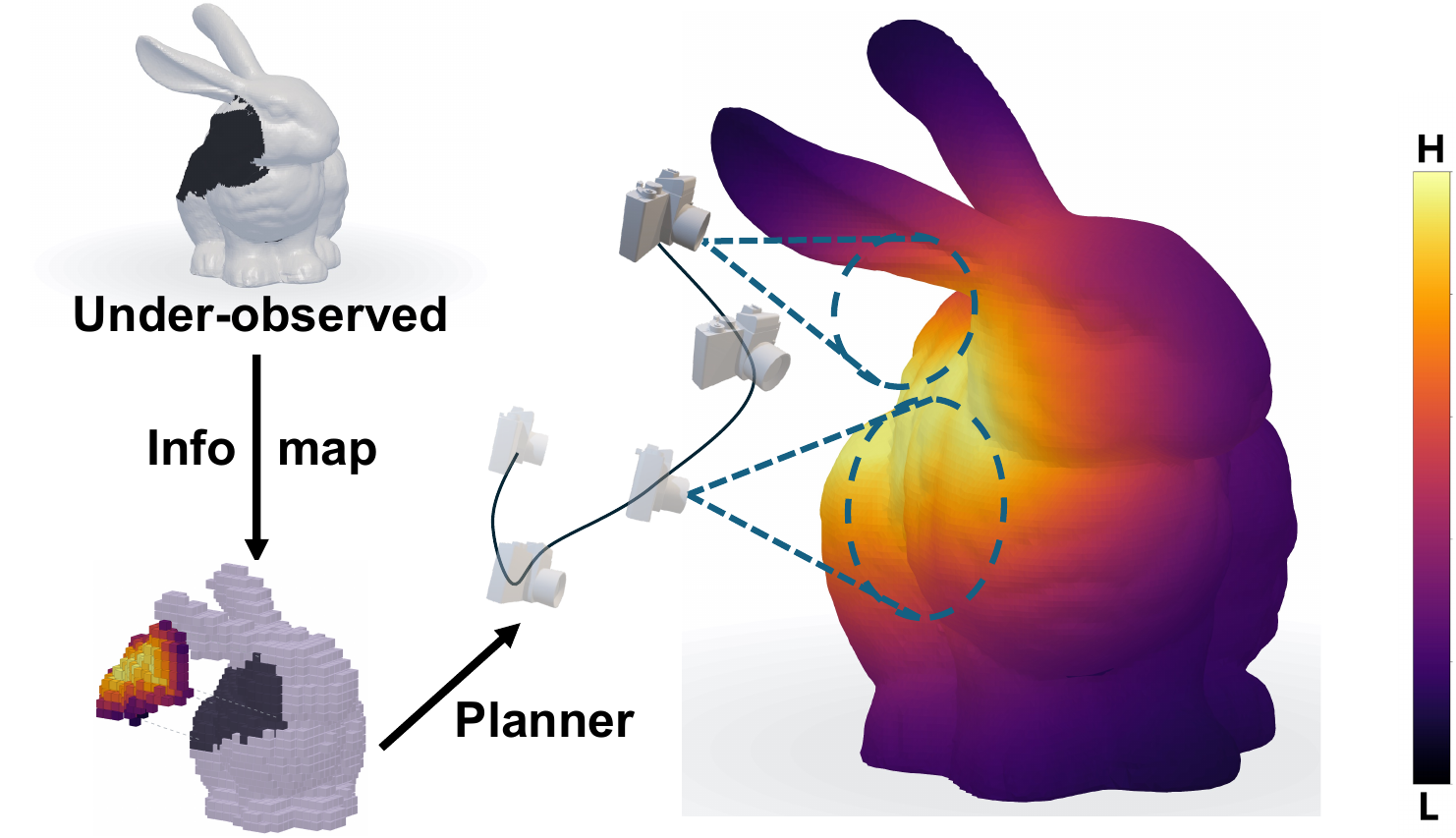}
\caption{\textbf{From reconstruction state to information distribution to ergodic coverage.}
The current reconstruction has well-built and under-built
regions (top left); the voxel map discretizes this state, and the
under-built region carries high information (bottom left); 
\textbf{Right:} the induced information distribution over the object
surface and a continuous trajectory whose view cones
dwell on high-information regions---the time spent looking at a region
is proportional to its information density.}
    \label{fig:ergo_search_object}

\end{figure}

We plan the trajectory with the kernel-ergodic metric of \citet{kenerl}, which
drives a trajectory's time-averaged positions to match a target distribution
$\phi$. Active reconstruction violates its central premise: the high-$\phi$ mass
lies on scene surfaces the robot cannot occupy, so no collision-free trajectory
can match it. We resolve this in two ways. First, we \emph{project} the target
into free space---the diffusion and masking of Sec.~\ref{sec:method-phi}, so the
body covers reachable space rather than the surfaces themselves. Second, we add two
footprint-aware mechanisms on top of the position-based metric: a
\emph{footprint-overlap depletion} term (Eq.~\eqref{eq:depletion}) that discourages
re-observing covered surfaces, and a \emph{footprint gaze reward}
(Eq.~\eqref{eq:gaze}) that aims the camera at high-$\phi$ surfaces while the body
stays in free space. Unlike \citet{zheng2025footprint}, which replaces the
point-sensor delta with a footprint distribution, our footprint enters only through
the depletion factor while the kernel-ergodic metric stays position-based, and we
derive $\phi$ online from a live 2DGS map rather than a fixed target.
Fig.~\ref{fig:ergo_search_object} illustrates the design on an object-centric
example.

  \paragraph{Time-varying $\phi$ via footprint-overlap depletion.}
  In active sensing, the marginal value of observing a region diminishes
  once prior visits have already covered it; $\phi$ should therefore
  decay at already-covered locations as the trajectory progresses. We model this decay by replacing the static evaluation
  $\phi_t(\mathbf{x}^k_t)$ with a coverage-discounted value
  
  \begin{equation}
      \phi^k_t \;\equiv\; \phi_t(\mathbf{x}^k_t) \prod_{j < k}
          \bigl(1 - \eta \cdot v_{jk}\bigr),
      \label{eq:depletion}
  \end{equation}
  where $\eta \in (0,1)$ is the per-visit discount rate and
  $v_{jk} \in [0,1]$ is a soft footprint-overlap kernel between waypoints
  $j, k$. We approximate the sensor footprint at waypoint $k$ by
  $D$ points $\mathbf{f}^k_{t,d} = \mathbf{x}^k_t + d\, \mathbf{z}^k_t$
  sampled along the optical axis $\mathbf{z}^k_t$ at depths
  $\{d_1, \dots, d_D\}$, each a candidate surface
  location at distance $d$ from the camera; the overlap kernel is then
  \begin{equation}
      v_{jk} \;=\; \max_{d_k,\, d_j} \,\exp\!\left(
          -\frac{\|\mathbf{f}^k_{t, d_k} - \mathbf{f}^j_{t, d_j}\|^2}
                {2\, \sigma_{\mathrm{fp}}^2}
      \right).
      \label{eq:footprint-overlap}
  \end{equation}
This serves as a differentiable proxy for surface co-visibility: two
waypoints whose central rays pass through nearby observable points are
penalized for re-observing the same region, with $\sigma_{\mathrm{fp}}$
acting as an effective tolerance that softens each ray into a cylinder.
The depletion product in Eq.~\eqref{eq:depletion} runs over the waypoints of the
current horizon; re-coverage \emph{across} horizons is suppressed separately, by
re-deriving $\phi_t$ from the updated map at every horizon boundary, where
well-built voxels drop out of Eq.~\eqref{eq:phi-raw}.

  \paragraph{Footprint-based information attraction.}
  The depletion-aware ergodic metric still evaluates $\phi$ at waypoint
 $\mathbf{x}^k_t$, so its gradient attracts the robot toward
  high-$\phi$ positions, which in reconstruction lie on or behind
  surfaces. We complement this with a footprint reward that attracts the
  camera's footprint to high-$\phi$ surfaces, while the robot position
  remains governed by the ergodic metric and safety penalties. 
  The reward averages the raw information map $\phi_t^{\mathrm{raw}}$
  (Eq.~\eqref{eq:phi-raw}) over the same footprint samples used by the
  depletion kernel, using the raw version to preserve absolute scale
  across horizons:
  \begin{equation}
      L_{\mathrm{gaze}}(\mathbf{u}_t) \;=\;
          -\frac{1}{K D}\sum_{k=1}^{K}\sum_{d=1}^{D}
              w_{k,d}\, \phi_t^{\mathrm{raw}}(\mathbf{f}^k_{t,d}),
      \label{eq:gaze}
  \end{equation}
where $w_{k,d} = \exp\bigl(-\kappa\, d_{\mathrm{unsafe}}(\mathbf{f}^k_{t,d})\bigr)$
  attenuates footprint samples in unobservable space. Here
  $d_{\mathrm{unsafe}}(\mathbf{f})$ is the distance from $\mathbf{f}$ to the
  observed collision-free region of $\mathcal{V}_t$: zero in free space and growing
  both inside obstacles \emph{and} in the unobserved region behind them, so the
  reward never credits pointing at high-$\phi$ voxels visible only through a wall.

\paragraph{Full cost and finite-horizon trajectory optimization.}
The depletion-aware ergodic metric is
  \begin{align}
      E_{\mathrm{kernel}}^{\mathrm{dep}}(\tau_t)
          \;=&\; -\frac{2}{K} \sum_{k=1}^{K} \phi^k_t 
            \; +\; \frac{1}{K^2} \\ \notag
            &\sum_{i,j=1}^{K}
              \exp\!\left(
                  -\frac{\|\mathbf{x}^i_t - \mathbf{x}^j_t\|^2}{2 \sigma^2}
              \right),
      \label{eq:kernel-ergodic-dep}
  \end{align}

where $\phi^k_t$ is the depletion-discounted target value of
Eq.~\eqref{eq:depletion} (information term) and the pairwise term repels nearby
waypoints toward uniform coverage (self-correlation term)---the two terms of
the kernel-ergodic metric of \citet{kenerl}, now evaluated on the
free-space-projected target of Sec.~\ref{sec:method-phi}. The box-filter diffusion
plays the mollifying role of the metric's Gaussian kernel, so evaluating the
diffused $\phi_t$ pointwise realizes the information term on a target the sensor can
reach: when the diffusion width equals the kernel bandwidth this is the metric of
\citet{kenerl} exactly, and otherwise a free-space-projected kernel-ergodic
objective.

\begin{figure*}[t]
    \centering
    \includegraphics[width=0.9\linewidth]{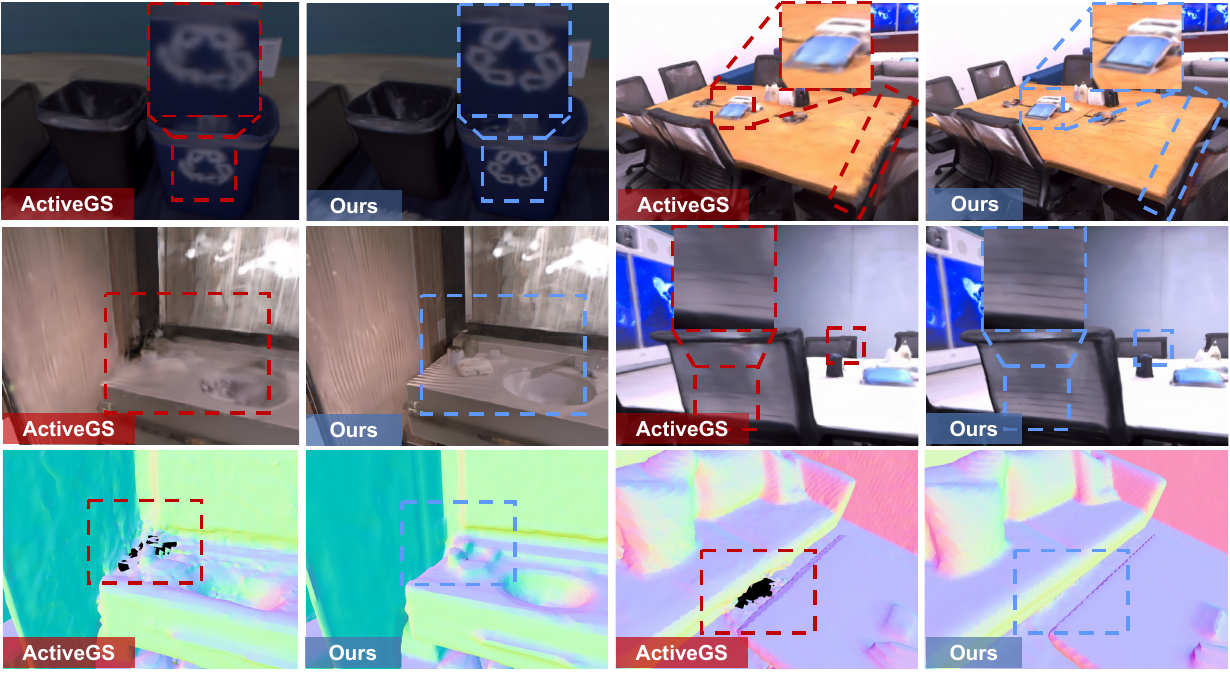}
    \caption{\textbf{Qualitative results of ActiveGS and our method on the Replica dataset.} We show RGB renderings (top two rows) and reconstructed surface meshes (bottom row) across four scenes. Red boxes on the ActiveGS results mark regions of low-quality reconstruction, and blue boxes mark the same regions in our results, which appear sharper and more faithful. By replacing greedy NBV selection with ergodic search over an information map, our method achieves higher-quality reconstruction.}
    \label{fig:vis}
\end{figure*}

  The full per-horizon cost combines the
  depletion-aware ergodic metric with the gaze reward, soft-barrier
  penalties $L_{\mathrm{safe}}$ for collisions, excessive step size, and
  out-of-bounds waypoints, and a quadratic control regularizer:
  \begin{equation}
\label{eq:full-cost}
\begin{split}
    J_t(\mathbf{u}_t) ={}& E_{\mathrm{kernel}}^{\mathrm{dep}}(\tau_t(\mathbf{u}_t))
        + \lambda_{\mathrm{g}}\, L_{\mathrm{gaze}}(\mathbf{u}_t) \\
    &+ \lambda_{\mathrm{s}}\, L_{\mathrm{safe}}(\mathbf{u}_t)
        + \lambda_{\mathrm{r}}\, \|\mathbf{u}_t\|^2.
\end{split}
\end{equation}

  All terms are differentiable in $\mathbf{u}_t$; we optimize with
  Adam~\citep{adam}, warm-starting from the previous-horizon
  solution. The agent executes all $K$ waypoints before re-planning.

\section{Experiments}
\subsection{Implementation Details}

\noindent\textbf{Dataset and simulator.}
  We evaluate \ourme on eight indoor scenes from the Replica
  dataset~\cite{replica}, following the active-reconstruction
  protocol of ActiveGS~\cite{jin2025activegs}. All experiments use the
  Habitat simulator~\cite{habitat} with an RGB-D camera moving freely in $\mathbb{R}^3$ with yaw and pitch control 
  , $[60^\circ \times 60^\circ]$ FOV,
  $512\!\times\!512$ resolution, $[0.1, 5.0]$~m depth range, and
  Gaussian depth noise $\sigma = 0.01d$. The world is discretized into
  a $20$~cm voxel grid. 

\noindent\textbf{Real Robot Experiments.}
We deploy the identical planner on two physical platforms: a \textbf{Unitree Go2}
quadruped that reconstructs a room-scale scene, and an\textbf{ Franka FR3 }arm that
reconstructs an object-centric scene. Maps are trained on real RGB-D data.
Sec.~\ref{sec:real-world} details the quadruped deployment; the manipulator
result is in the supplementary material.

  \noindent\textbf{Training and planning schedule.}
  All experiments run on a single RTX 5090.
  Each mission runs for a time budget of $300$~s, accumulated
  as $t_\text{map} + t_\text{plan} + t_\text{fly}$, where $t_\text{fly}$ is calculated from the executed path length divided by a constant velocity of $1$~m/s. Per horizon, the planner optimizes a $10$-step trajectory with
  Adam ($40$ iterations) warm-started from the previous solution;
  after execution, the Gaussian map is trained for $12$ gradient steps
  on a mini-batch of $10$ frames sampled along the executed path
  together with $10$ frames drawn from the observation history.

   \noindent\textbf{Evaluation metrics.}
  We report PSNR, SSIM, and
  LPIPS on a fixed set of $1000$ test
  viewpoints sampled uniformly in each scene's free space,\textbf{ averaged
over 5 independent runs}. Geometry
  is examined qualitatively in Sec.~\ref{sec:qualitative}. All
  evaluation settings match primary
  baseline~\cite{jin2025activegs}.

  \noindent\textbf{Baselines.}
  We compare against ActiveGS~\citep{jin2025activegs}, an NBV planner
  re-run in our pipeline under matched mapper, budget, and evaluation
  protocol (same as their setting). For completeness, Table~\ref{tab:replica_comparison} also reports NARUTO~\citep{feng2024naruto},
  and FisherRF~\citep{jiang2024fisherrf} with numbers cited from
  \citet{jin2025activegs}, which already established ActiveGS as the strongest of these earlier methods. Sampling-density ablations are described in Sec.~\ref{sec:ablation}.

  \subsection{Quantitative Analysis}
  \label{sec:comparison}
  \ourme\ achieves higher PSNR than ActiveGS on all scenes (Table~\ref{tab:replica_comparison}),
  with a mean gain of \textbf{$\mathbf{1.5}$~dB}; LPIPS and SSIM show the same pattern. The lead arises from two mechanisms: continuous-pose
  optimization places the camera at any high-$\phi$ pose along the trajectory rather than the nearest discrete candidate, and
  depletion suppresses re-coverage, spreading subsequent waypoints over low-confidence surfaces.
The PSNR gain is positive on every scene, but its composition differs by scene type. On the office scenes, SSIM is comparable, while LPIPS improves by \textbf{$\mathbf{6}$--$\mathbf{19\%}$}, indicating that the gain is concentrated in fine-scale appearance. We attribute this to continuous-pose optimization, which reaches poses unavailable to discrete candidates. On the room and hotel scenes, the SSIM gap widens to \textbf{$\mathbf{0.012}$--$\mathbf{0.017}$}, indicating that structural differences also emerge. These scenes contain heavier occlusion from furniture, where surfaces missed by discrete viewpoint selection require deliberate coverage, which the ergodic trajectory provides by continuously varying its heading. 
The gain is not explained by frame count alone:
Sec.~\ref{sec:ablation} shows that\textbf{ densifying the baseline's sampling
along its path does not close the gap.}

  \subsection{Qualitative Results}
  \label{sec:qualitative}

  Figure~\ref{fig:vis} compares reconstructions from \ourme\ and
  ActiveGS on Replica scenes. The most pronounced differences
  lie in surface fidelity: in the
  highlighted regions, \ourme\ (blue boxes) recovers sharper
  texture, consistent lighting, and finer geometric detail on
  furniture and wall surfaces, while ActiveGS (red boxes) produces
  flatter, lower-fidelity reconstructions on the same regions. These
  fidelity gains arise from the same trajectory-level coupling that
  drives the PSNR lead: depletion keeps the camera looking at
  under-confident surfaces rather than re-sampling already-covered
  ones. 
  The regions that differ visually are also where LPIPS separates most,
confirming that the gain is concentrated in high-frequency texture rather
than spread uniformly over the image.
The mesh row shows the same pattern, where ActiveGS leaves holes on
surfaces its path merely passes, while
\ourme{} recovers a continuous surface in the same regions.

\begin{table*}[th!]
\centering
\caption{\textbf{Ablation study on Replica (PSNR).} \baselineAR and \baselineAU
observe at $10$ random / uniformly-interpolated poses along ActiveGS's path;
\emph{Ours-Kernel ES} removes the footprint mechanism of
Sec.~\ref{sec:method-ergodic}.}
\label{tab:ablation}
\begin{tabularx}{\textwidth}{l*{8}{C}}
\toprule
\textbf{Method} & \textbf{Of0} & \textbf{Of2} & \textbf{Of3} & \textbf{Of4} & \textbf{R0} & \textbf{R1} & \textbf{R2} & \textbf{H0} \\
\midrule
ActiveGS       & \second{36.78} & \second{31.68} & \third{32.16} & {34.08}  & \third{29.93} & \second{31.74} & {32.35}  & \third{32.04} \\
\baselineAR    & \third{36.70}  & \third{31.37}  & \second{32.70}  & \second{34.35} & \second{30.09}  & \third{31.29}  & \third{32.79} & \second{32.11}  \\
\baselineAU    &       35.86    &       29.98    &       30.49    &       32.38    &       27.63    &       30.95    &       31.65    &       28.24    \\
Ours-Kernel ES &       35.63    &       30.78    &       31.83    &       \third{34.09}    &       29.91    &       31.15    &       \second{33.04}    &       29.39    \\
\textbf{Ours}  & \first{38.36}  & \first{32.81}  & \first{33.79}  & \first{34.92}  & \first{31.55}  & \first{32.84}  & \first{34.70}  & \first{33.69}  \\
\bottomrule
\end{tabularx}
\end{table*}

\subsection{Ablation Study}
  \label{sec:ablation}
To examine whether \ourme's gain simply comes from denser sampling along the trajectory, and to validate the effectiveness of footprint depletion, we design the following ablations.
For the first, we implement two ActiveGS variants:
\baselineAR scatters $10$ random observations per horizon (a baseline without planner), and \baselineAU captures $10$ uniformly-interpolated poses along its original path (the information reachable from the trajectory geometry alone). Table~\ref{tab:ablation} reports that \baselineAR matches ActiveGS within noise \textbf{($\mathbf{+0.08}$~dB)}, and \baselineAU drops \textbf{($\mathbf{-1.7}$~dB)}. Neither closes the gap to \ourme. 
\textbf{Simply densifying observations performs even worse, as redundant views dilute the training weight of informative ones.}
For the second, we drop the footprint mechanism in Sec.~\ref{sec:method-ergodic}, leaving the original kernel-ergodic metric from \citet{kenerl}:
This variant loses \textbf{$\mathbf{2.1}$~dB} and falls below ActiveGS on six of eight
scenes. 
NBV replans after every view, and its confidence updates implicitly deplete visited regions. Ergodic search alone optimizes the whole horizon against a frozen information map, yielding redundant views. Depletion is therefore essential to our framework.

\subsection{Executing Trajectories in Real World}
\label{sec:real-world}
We deploy \ourme{} on two physical platforms: \textbf{a quadruped (Unitree Go2)} exploring a room-scale scene with the same planner as in Sec.~\ref{sec:method-ergodic} and a \textbf{Franka arm for object reconstruction (supplementary materials)}. Since the planned
trajectory is itself the optimization variable, its waypoints are
handed to each platform's controller directly, with no intermediate
path planner. This is a practical payoff of trajectory-level planning.
An NBV planner commits to discrete viewpoints and delegates the
connecting motion to a separate path planner, which can issue \textbf{dynamically infeasible} commands on a physical platform.
\ourme{} instead optimizes the trajectory under the platform's own
dynamics, so every planned waypoint is
\textbf{dynamically consistent by construction}. As shown in the supplementary materials, \ourme{} achieved a $\mathbf{100 \%}$ \textbf{success rate}, whereas our NBV baseline made minor contact with the environment in every trial.

\begin{figure}[ht]
    \centering
    \includegraphics[width=0.95\linewidth]{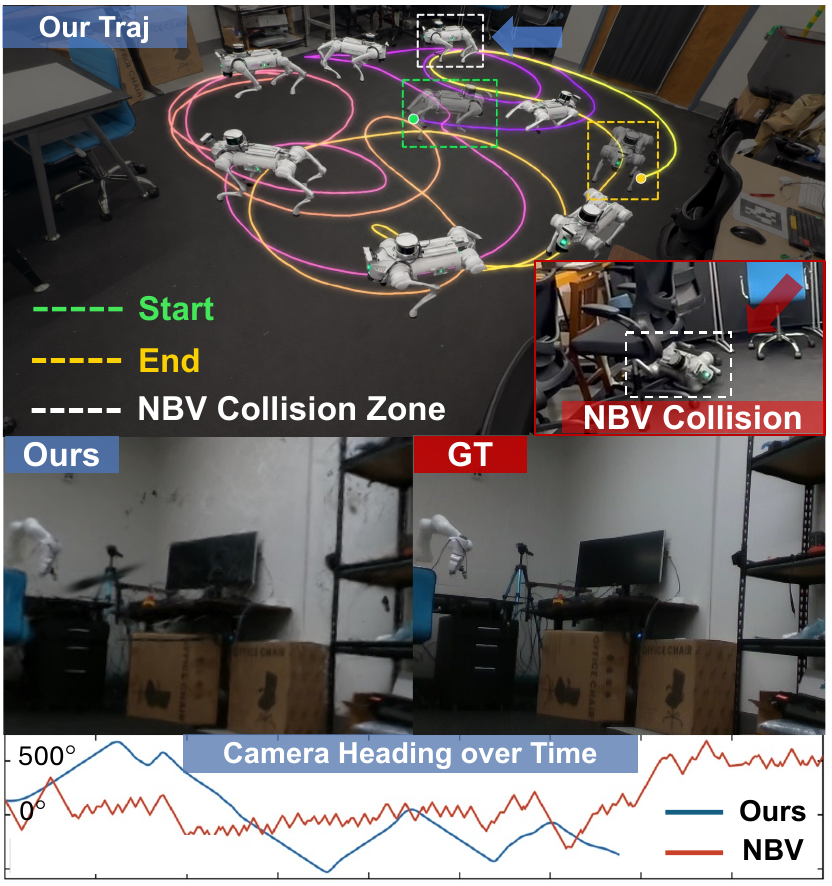}
\caption{{ \textbf{\ourme{} on Unitree Go2.}
\textbf{Top:} the Unitree Go2 executes a planned ergodic trajectory
in a laboratory scene (composite with time-colored path); the dashed
boxes mark the start and final poses and a region that the NBV
baseline could not traverse safely (inset). \textbf{Middle:} Visualization of the reconstruction quality. \textbf{Bottom:} the ergodic trajectory turns smoothly and continuously, whereas NBV produces rapid heading reversals between
committed viewpoints.}}
    \label{fig:RealRobot}

\end{figure}

\paragraph{Room-scale exploration (Unitree Go2).}
A quadruped carries a front-mounted RGB-D camera (Intel D435). The
trajectory is constrained to the traversable plane with yaw aligned to the
direction of motion, and waypoints are tracked by a velocity-level controller
built on the manufacturer's high-level interface, without an intermediate path
planner. Over $5$-minute missions in a $42$\,m$^2$ scene, every planned horizon
executes without infeasible commands, and observations are integrated into the
2DGS map. In Fig.~\ref{fig:RealRobot}, the time-colored ergodic path~(top)
sweeps the room as a single continuous trajectory; the dashed inset marks a
cluttered region the \textbf{NBV baseline cannot traverse without collision, while
\ourme{} threads the free space safely.} The reconstruction~(middle) is clean and
complete. The heading traces~(bottom) show our heading varying continuously,
whereas NBV snaps between discrete targets with rapid reversals, exactly the
motion a viewpoint-level planner cannot penalize.

\section{Conclusion}
\label{sec:conclusion}
We presented \ourme, an ergodic-trajectory formulation for active
2D Gaussian-Surfel reconstruction. Instead of committing to discrete
next-best-views, \ourme{} derives a target information distribution
online from voxel-level information and per-Gaussian uncertainty, and optimizes a
continuous, control-feasible trajectory with a kernel-ergodic horizon
planner and footprint-overlap depletion. Across all eight Replica
scenes, \ourme{} improves over our baseline (ActiveGS) by
$+1.5$~dB PSNR on average, and its trajectories execute directly on
real physical robot without an intermediate path planner. By recasting active reconstruction as ergodic coverage, \ourme{}
plans information gathering and feasible motion jointly within a
single objective.

\section{Acknowledgments}
This work was also supported in part by funding from the Johns Hopkins Data Science
and AI Institute. Thanks Rex for insightful discussion on our physical experiments.
\bibliography{aaai2027}

\appendix

\appendix

\twocolumn[{%
  \centering
  {\LARGE\bfseries TRACE: Ergodic Trajectory Optimization for Active Scene Reconstruction Supplementary Material\par}
  \vspace{2em}
}]

\section{Per-Horizon Algorithm}
\label{app:algorithm}
Algorithm~\ref{alg:etosplat} summarizes one plan-and-execute horizon of
\ourme{}; the notation follows Sec.~\ref{sec:method}.

\begin{algorithm}[h]
\caption{\ourme{} per-horizon plan-and-execute}
\label{alg:etosplat}
\KwIn{Live 2DGS map $\mathcal{M}_t$, voxel map $\mathcal{V}_t$, current pose
      $\mathbf{p}_t$, warm-start $\mathbf{u}_{\mathrm{prev}}$}
$\phi_t \gets \textsc{BuildInfoMap}(\mathcal{M}_t,\mathcal{V}_t)$
  \tcp*{\eqref{eq:phi-raw}}
$\mathbf{u} \gets \textsc{WarmStart}(\mathbf{u}_{\mathrm{prev}})$\;
\For{$i = 1,\dots,N_{\mathrm{iter}}$}{
  $\tau \gets \textsc{Rollout}(\mathbf{p}_t,\mathbf{u})$\;
  Compute $J(\mathbf{u})$ via Eqs.~\eqref{eq:depletion}--\eqref{eq:full-cost}\;
  $\mathbf{u} \gets \textsc{AdamStep}(\mathbf{u},\nabla_{\mathbf{u}}J)$\;
}
$\tau_t \gets \textsc{Rollout}(\mathbf{p}_t,\mathbf{u})$\;
Execute $\tau_t$; integrate RGB-D into $\mathcal{M}_{t+1},\mathcal{V}_{t+1}$\;
$\mathbf{u}_{\mathrm{prev}} \gets \mathbf{u}$\;
\end{algorithm}
\section{Joint-Space Ergodic Search on the FR3}
\label{app:franka}
On the Franka FR3 arm, we plan directly in joint space. A single integrator sufficed
for the Go2 because it has a velocity interface; the arm does not. The
trajectory variable is instead a sequence of joint configurations, and a
differentiable forward-kinematics chain maps each to the camera pose, which
determines where the world-frame information map $\phi$ is sampled. This makes
reachability intrinsic: with joint limits as box constraints, every planned
view is reachable, and the plan runs with \textbf{no inverse kinematics, no candidate
viewpoint set, and no feasibility repair.} Collisions with the table, wall,
and base column are penalized in the same objective, on forward-kinematic body
points. Here the ergodic objective of Sec.~\ref{sec:method-ergodic} is optimized
through the nonlinear forward kinematics rather than over camera poses directly,
yet every executed waypoint is dynamically consistent.

\paragraph{Setup.}
The arm's state is a joint configuration $q\in\mathcal{Q}\subset\mathbb{R}^{7}$.
Working within a single horizon, we drop the horizon index $t$; the planner
optimizes a $K$-step joint trajectory
$\mathbf{q}=(q^{1},\dots,q^{K})$ from joint-velocity controls $\mathbf{u}$, with
$q^{k}=q^{k-1}+\Delta t\,u^{k}$. The camera is mounted on the end-effector, so
its pose follows from the configuration through forward kinematics.

\begin{definition}[Kinematic Sensor Map]\label{def:fk}
Let $\mathrm{FK}:\mathcal{Q}\to SE(3)$ be the forward-kinematics map to the
camera pose, $\mathrm{FK}(q)=T_{\mathrm{ee}}(q)\,T_{\mathrm{cam}}$, with
$T_{\mathrm{cam}}$ the fixed hand--eye transform. We write
$\mathrm{FK}(q)=(R(q),\mathbf{x}(q))$, with $R(q)\in SO(3)$ the camera
orientation and $\mathbf{x}(q)\in\mathbb{R}^{3}$ its position.
\end{definition}

A configuration fixes both where the camera sits, through $\mathbf{x}(q)$, and
where it looks, through the optical axis $\mathbf{z}(q)=R(q)\mathbf{e}_{3}$. We model its view as samples along this axis.

\begin{figure}[htb!]
    \centering
       \includegraphics[width=0.9\linewidth]{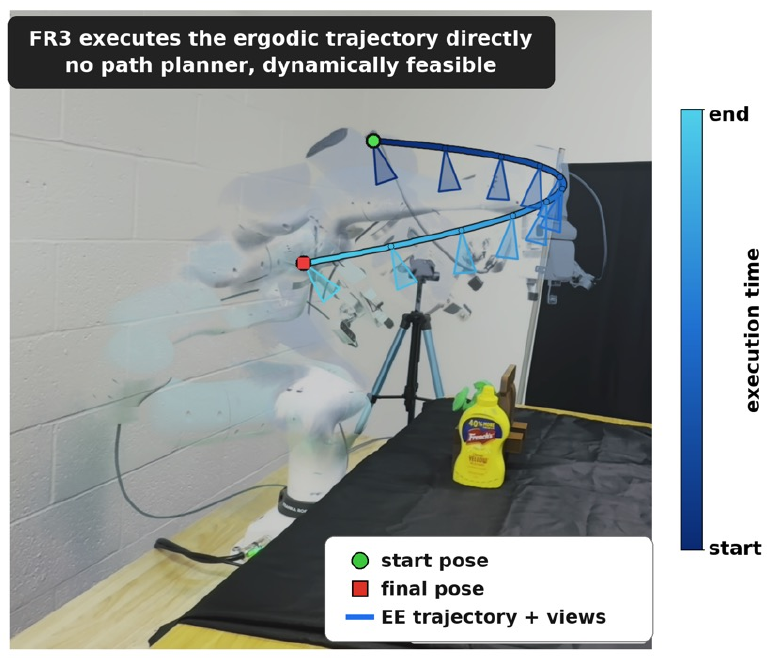}
    \caption{\textbf{\ourme{} on a Franka FR3.} The arm executes the planned
ergodic trajectory directly, with no path planner. We overlay the executed
end-effector (EE) path, its camera frustums, and the
start and final poses (green circle, red square). Planning in joint space makes
every waypoint dynamically consistent. \textbf{The supplementary video shows this in full.}}
    \label{fig:franka}
\end{figure}
\begin{definition}[Sensor Footprint]\label{def:footprint}
The footprint at $q$ is the set of world points
\begin{equation}\label{eq:footprint}
\mathbf{f}_{d}(q)=\mathbf{x}(q)+d\,\mathbf{z}(q),\qquad d\in\{d_1,\dots,d_D\},
\end{equation}
sampled along the optical axis at depths $d_1,\dots,d_D$.
\end{definition}

\begin{figure*}[t!]
    \centering
    \includegraphics[width=.85\linewidth]{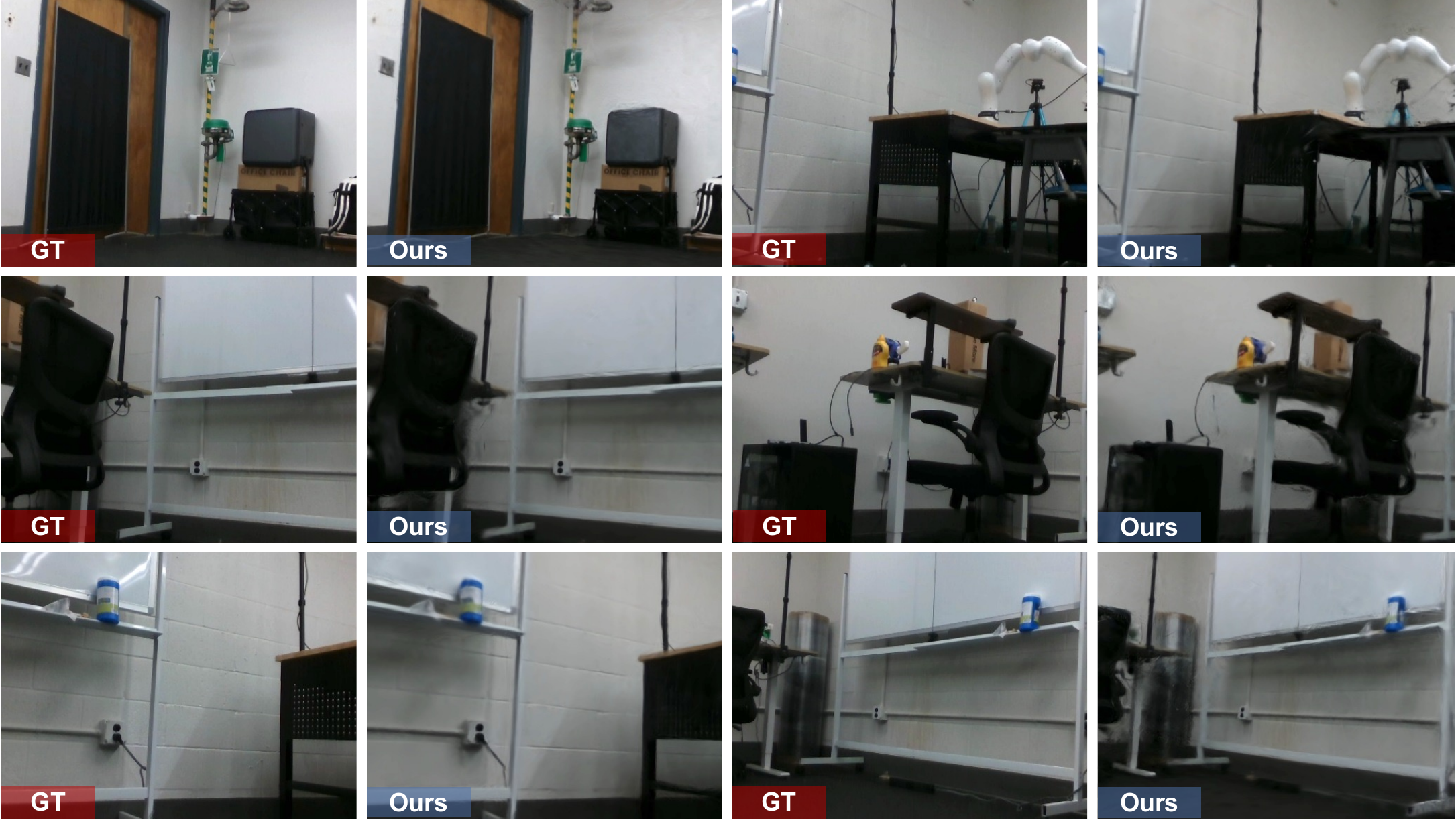}
    \caption{\textbf{Real scene versus \ourme{} reconstruction on the Unitree Go2.} The room-scale reconstruction, produced from the directly executed ergodic trajectory, recovers the scene geometry and appearance.\textbf{The
supplementary video shows this in full.}}
    \label{fig:go2recon}
\end{figure*}
\paragraph{Objective.}
We keep the position-space kernel-ergodic metric of \citet{kenerl}, which scores a trajectory
by how closely its time-averaged occupancy matches the target distribution and
splits into an information and a uniform-coverage term. Evaluated on the
camera positions $\mathbf{x}(q^{k})$, it reads
\begin{align}
\label{eq:kernel-joint}
E_{\mathrm{kernel}}^{\mathrm{dep}}(\mathbf{q})=&-\frac{2}{K}\sum_{k=1}^{K}\phi^{k}
+\frac{1}{K^{2}} \\ \notag
&\sum_{i,j=1}^{K}\exp\!\left(
-\frac{\|\mathbf{x}(q^{i})-\mathbf{x}(q^{j})\|^{2}}{2\sigma^{2}}\right).
\end{align}
The first term is the information term: it evaluates the depletion-discounted
map $\phi^{k}=\phi(\mathbf{x}(q^{k}))\prod_{j<k}(1-\eta\cdot v_{jk})$ at each camera
position and pulls the trajectory toward informative regions, where $v_{jk}$ is
the footprint-overlap kernel
(Eqs.~\eqref{eq:depletion}--\eqref{eq:footprint-overlap}). The second is the
coverage term: a Gaussian kernel between camera positions that penalizes
revisiting and spreads the trajectory out. They reproduce the
information-and-coverage decomposition of \citet{kenerl}, now on the camera positions the configuration produces.

Position alone does not fix what the camera sees, so orientation enters through
the footprint. The gaze term evaluates the raw map $\phi^{\mathrm{raw}}$ over the
footprint samples,
\begin{equation}\label{eq:gaze-joint}
L_{\mathrm{gaze}}(\mathbf{q})=-\frac{1}{KD}\sum_{k=1}^{K}\sum_{d=1}^{D}
w_{k,d}\,\phi^{\mathrm{raw}}\!\big(\mathbf{f}_{d}(q^{k})\big),
\end{equation}
with $w_{k,d}$ down-weighting occluded samples; its gradient turns $R(q)$ toward
high-information surfaces. Position and orientation are therefore optimized
together, both as functions of the single map $\mathrm{FK}$.

\begin{problem}[Joint-Space Ergodic Search]\label{prob:joint}
Given an information map $\phi$, minimize the objective over joint-velocity
controls subject to the arm's kinematics:
\begin{align*}
\min_{\mathbf{u}}\ \;&E_{\mathrm{kernel}}^{\mathrm{dep}}(\mathbf{q})+\lambda_{\mathrm{g}}L_{\mathrm{gaze}}(\mathbf{q})
+\lambda_{\mathrm{s}}L_{\mathrm{safe}}(\mathbf{q})+\lambda_{\mathrm{r}}\|\mathbf{u}\|^{2}\\
\text{s.t.}\ \;&q^{k}=q^{k-1}+\Delta t\,u^{k},\qquad q^{k}\in[q_{\min},q_{\max}],
\end{align*}
where $L_{\mathrm{safe}}$ is the soft-barrier penalty of the full cost
\eqref{eq:full-cost}, evaluated on forward-kinematic body points.
\end{problem}

\begin{proposition}[Feasibility and Generality]\label{prop:joint}
The joint-space problem has three properties. First, the objective is
differentiable in $\mathbf{u}$: the forward kinematics is differentiable and the
kernel-ergodic and gaze terms are differentiable in the camera pose, so the arm
reuses the quadruped's gradient-based optimizer unchanged. Second, every
trajectory that respects the joint limits $q^{k}\in[q_{\min},q_{\max}]$
maps to an achievable camera pose $\mathrm{FK}(q^{k})$, so every planned waypoint
is executable without inverse kinematics or feasibility repair. Third, when the
decision variable is the camera pose itself, the forward-kinematics map drops
out and the problem reduces to the camera-pose objective of the main paper,
recovering the quadruped case.
\end{proposition}

 \begin{table}
  \centering
  \caption{\textbf{Executing ActiveGS viewpoints directly on the FR3
  (simulation).} Across $10$ paired trials with diverse start
  poses under a shared mapper and budget, we command each planner's output on
  the arm without a path planner. Every ActiveGS trial collides with the scene
  and knocks the object off the table; \ourme{} stays collision-free.} 
  \label{tab:franka-collision}
  \setlength{\tabcolsep}{6pt}
  \renewcommand{\arraystretch}{1.0}
  \begin{tabular}{lc}
    \toprule
    \textbf{Method} & \textbf{Collision-free trials} \\
    \midrule
    ActiveGS        & $0/10$ \\
    \first{\textbf{\ourme{} (Ours)}} & \first{$\mathbf{10/10}$} \\
    \bottomrule
  \end{tabular}
\end{table}

Figure~\ref{fig:franka} shows a representative mission on the real arm.
\ourme{} follows the joint-space trajectory in one continuous sweep around the
object, and every view executes without collision or intervention.
Table~\ref{tab:franka-collision} quantifies the contrast in simulation, where an
unrealizable plan can be run safely. Across $10$ paired trials under a shared
mapper and budget, the camera poses ActiveGS selects \textbf{collide with the scene on
every trial} and knock the object off the table, while \ourme{}, planning in
joint space, \textbf{stays collision-free throughout.}

\section{Room-Scale Reconstruction on the Go2}
\label{app:go2}
Figure~\ref{fig:go2recon} compares the reconstruction \ourme{} produces on the
Unitree Go2 with photo of the same scene. Over a
$5$-minute mission the quadruped executes the planned ergodic trajectory
directly, and the online 2DGS map fills in furniture and wall surfaces across
the room. The reconstruction is clean and complete, recovering geometry and
texture on the surfaces the trajectory sweeps, which confirms the executed path
is both \textbf{feasible and information-rich at room scale.}

\end{document}